\documentclass[10pt,twocolumn,letterpaper]{article}

\usepackage[pagenumbers]{cvpr}     
\usepackage{graphicx}
\usepackage{amsmath}
\usepackage{amssymb}
\usepackage{booktabs}

\usepackage{algorithm}
\usepackage{algorithmic}
\usepackage{multirow}
\usepackage{bbding}

\usepackage{enumitem}

\usepackage{color, colortbl}
\usepackage[dvipsnames]{xcolor}
\usepackage[pagebackref,breaklinks,colorlinks]{hyperref}

\usepackage{multirow}

\newcommand{\hgreen}[1]{\textcolor{ForestGreen}{#1}} 

\definecolor{tabhighlight}{HTML}{e5e5e5}

\usepackage[capitalize]{cleveref}
\crefname{section}{Sec.}{Secs.}
\Crefname{section}{Section}{Sections}
\Crefname{table}{Table}{Tables}
\crefname{table}{Tab.}{Tabs.}

\newcommand*\samethanks[1][\value{footnote}]{\footnotemark[#1]}

\begin{document}


\title{Training-Free Unsupervised Prompt for Vision-Language Models}

\author{Sifan Long\textsuperscript{\rm 1,2}
\thanks{Equal contribution. The work was done during an internship at Baidu.}
~~~~Linbin Wang\textsuperscript{\rm 1}\samethanks
~~~~Zhen Zhao\textsuperscript{\rm 3}\samethanks
~~~~Zichang Tan\textsuperscript{\rm 2} \\
~~~~Yiming Wu\textsuperscript{\rm 4}
~~~~Shengsheng Wang\textsuperscript{\rm 1}\thanks{Corresponding authors.}
~~~~Jingdong Wang\textsuperscript{\rm 2} \\
\textsuperscript{\rm 1}Jilin University\hspace{6mm}
\textsuperscript{\rm 2}Baidu VIS\hspace{6mm}
\textsuperscript{\rm 3}University of Sydney\hspace{6mm}
\textsuperscript{\rm 4}University of Hong Kong\\
{\tt\small \{longsf22, wanglb23\}@mails.jlu.edu.cn~~zhen.zhao@sydney.edu.au
~~yimingwu0@gmail.com}\\
{\tt\small wss@jlu.edu.cn~~\{tanzichang, wangjingdong\}@baidu.com}\\
}

\maketitle 
\begin{abstract}
Prompt learning has become the most effective paradigm for adapting large pre-trained vision-language models to downstream tasks.
Recently, unsupervised prompt tuning methods, such as UPL and POUF, directly leverage pseudo-labels as supervisory information to fine-tune additional adaptation modules on unlabeled data.
However, inaccurate pseudo-labels easily misguide the tuning process and result in poor representation capabilities.
In light of this, we propose Training-Free Unsupervised Prompts (TFUP), which maximally preserves the inherent representation capabilities and enhances them with a residual connection to similarity-based prediction probabilities in a training-free and labeling-free manner.
Specifically, we integrate both instance confidence and prototype scores to select representative samples, which are used to customize a reliable Feature Cache Model (FCM) for training-free inference.
Then, we design a Multi-level Similarity Measure (MSM) that considers both feature-level and semantic-level similarities to calculate the distance between each test image and the cached sample as the weight of the corresponding cached label to generate similarity-based prediction probabilities.
In this way, TFUP achieves surprising performance, even surpassing the training-base method on multiple classification datasets.
Based on our TFUP, we propose a training-based approach (TFUP-T) to further boost the adaptation performance. 
In addition to the standard cross-entropy loss, TFUP-T adopts an additional marginal distribution entropy loss to constrain the model from a global perspective.
Our TFUP-T achieves new state-of-the-art classification performance compared to unsupervised and few-shot adaptation approaches on multiple benchmarks.
In particular, TFUP-T improves the classification accuracy of POUF by 3.3\% on the most challenging Domain-Net dataset~\footnote{Code and logs: \url{https://github.com/wlb12345/TFUP}}. 

\end{abstract}

\begin{figure}
	\centering
	\includegraphics[width=0.48\textwidth]{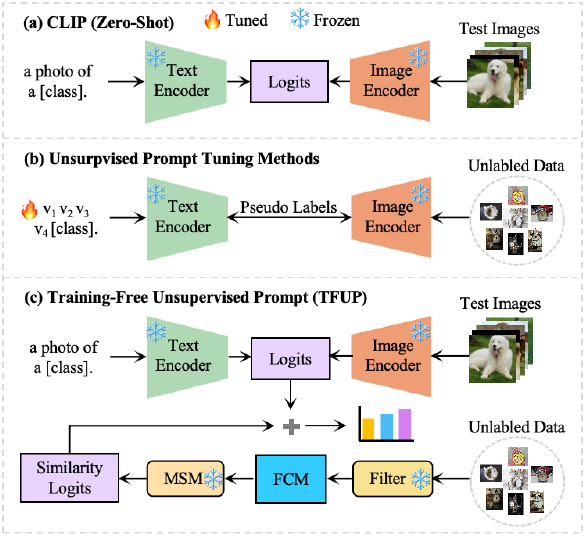}
	\caption{(a) Zero-shot inference of the pre-trained CLIP. (b) Existing unsupervised prompt tuning methods such as UPL \cite{huang2022unsupervised} and POUF \cite{tanwisuth2023pouf}, which fine-tune models or prompts directly on unlabeled data. (c) Our training-free unsupervised prompt (TFUP) method generates similarity-base prediction probabilities by customizing the proposed Feature Cache Model (FCM) and Multi-level Similarity Measure (MSM).}
	\label{FIG:1}
\end{figure}

\section{Introduction}
\label{sec:intro}

Foundational large-scale vision-language models (VLMs), such as CLIP \cite{radford2021learning} and ALIGN \cite{jia2021scaling}, have demonstrated impressive representation and generalization on various downstream tasks by using the contrastive learning objective on hundreds of millions of text and image pairs.
Due to a potential shift between the pre-training and the specific task \cite{liu2023graphprompt}, fine-tuning is a common method to bridge this gap, which leverages labeled data on downstream tasks to fine-tune all parameters of the pre-trained model~\cite{devlin2018bert}.
However, it requires significant amounts of labeled data and is computationally expensive, even leading to over-fitting \cite{houlsby2019parameter}.
To this end, prompt learning, as a parameter-efficient fine-tuning paradigm, recently attracted increasing attention from the research community.
As a representative work, CLIP \cite{radford2021learning} directly utilizes hand-crafted prompts to achieve impressive zero-shot classification performance in Fig. \ref{FIG:1} (a).
But the amount of manual prior knowledge required to design appropriate prompts for each specific task is intolerable.
Inspired by tuning studies in large language models (LLMs) ~\cite{li2021prefix,lester2021power}, latter studies like CoOp \cite{zhou2022learning} and CLIP-Adapter ~\cite{gao2021clip}, train learnable prompts on labeled data to alleviate such reliance on hard-prompt designs.

Though few-shot prompt methods gain significant improvements, they still require artificial prior knowledge to label downstream data and rely on manual annotation quality, which may limit the scalability of the original model. 
To this end, recent unsupervised prompt tuning frameworks have been introduced to eliminate the need for data annotations and enhance the efficiency of adapting VLMs for various downstream tasks~\cite{huang2022unsupervised,tanwisuth2023pouf}. 
As shown in Fig. \ref{FIG:1} (b), these methods tend to fine-tune models or learnable prompts directly on unlabeled data. 
UPL \cite{huang2022unsupervised} selects the top-K confidence samples per class to tune the whole model using pseudo labels generated by the pre-trained vision-language model. 
POUF \cite{tanwisuth2023pouf} treats the representation of class-specific text prompts as class prototypes and further aligns these prototypes with target image features in the latent space.
However, these methods often undervalue the pre-trained CLIP generalization ability, which is crucial for achieving robust performance across diverse downstream datasets. 
By focusing on optimizing the CLIP performance on a specific set, they may overlook the importance of learning general features that can be applied to new, unseen data~\cite{zhou2022conditional,yao2023visual}. 
Besides, these methods rely heavily on the pseudo labels to tune additional adaptation modules, which will inevitably introduce the confirmation bias~\cite{bias2019}. 
Inaccurate pseudo-labels can misguide the tuning process and result in poor generalization capabilities.
Consequently, our primary goal is to maximize the retention of the pre-trained VLMs' capabilities while adapting them to downstream tasks with minimal costs.

Motivated by above observed limitations, we propose Training-Free Unsupervised Prompt (TFUP), which maximally preserves the inherent representation capabilities and enhances them with a residual connection to similarity-based prediction probabilities in a training-free and labeling-free manner.
Our TFUP generates similarity-base prediction probabilities by customizing a Feature Cache Model (FCM) and designing a Multi-level Similarity Measure (MSM). 
As shown in Fig. \ref{FIG:1} (c), we first extract image features of unlabeled training images by CLIP's visual encoder and then calculate the cosine similarity between image features and text features to obtain the predicted probability of the current image.
In FCM, we select top-K samples as high-confidence samples for each category. Nonetheless, it is inevitable that these high-confidence images contain noisy information, such as complex backgrounds. In light of this, we further propose a prototype filter, which introduces an attention mechanism to filter out the representative samples from the constructed 
high-confidence sample set. Consequently, we can create a cache model using the features of the representative sample and the corresponding one-hot labels as key-value pairs.

\begin{figure}
	\centering
	\includegraphics[width=0.45\textwidth]{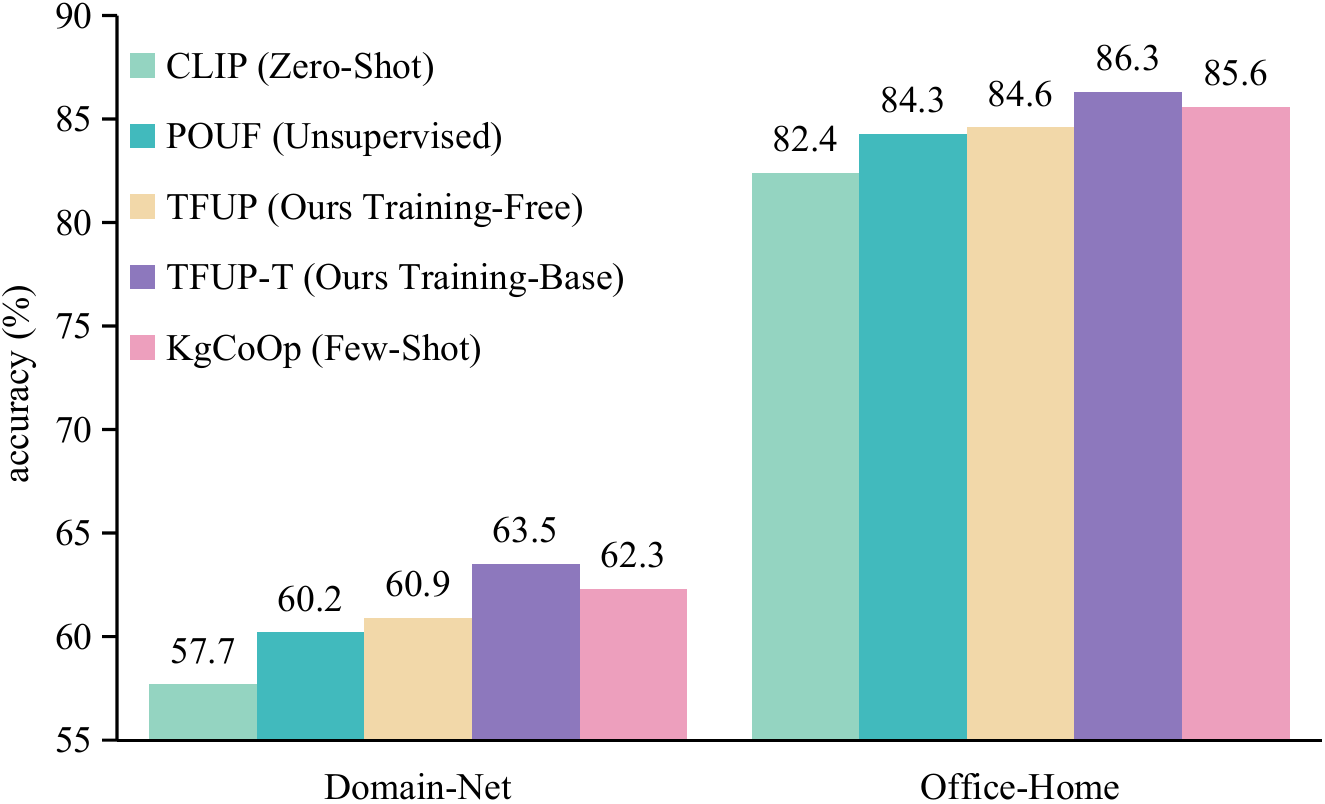}
    \caption{
    Performance comparisons of CLIP \cite{radford2021learning}, POUF \cite{tanwisuth2023pouf}, TFUP, TFUP-T, and KgCoOp \cite{yao2023visual} on Domain-Net and Office-Home datasets in terms of top-1 classification accuracy.
    }
	\label{FIG:2}
 \vspace{-0.2cm}
\end{figure}

Based on the constructed FCM, we calculate the distance between the test image and cached samples as the weights of the corresponding labels. Then, we combine different sample labels into a similarity-based prediction probability according to the weights.
In existing unsupervised tuning studies, the distance measure method only considers the similarity of the overall image information and easily introduces background noise. For example, two pictures with the same background but different foregrounds are judged to be of the same category. Differently, we design a new measure method called Multi-level Similarity Measure (MSM), which considers both feature-level and semantic-level similarities. 
Specifically, we not only calculate the cosine similarity between image features as feature-level similarity but also calculate the KL divergence between image prediction probabilities as semantic-level similarity. Ultimately, we leverage a hadamard product to combine feature and semantic similarities.
As shown in Fig. \ref{FIG:2}, through this non-parametric and non-training approach, our TFUP demonstrates extremely excellent efficiency and achieves promising performance, even surpassing the training-base unsupervised prompt learning methods \cite{huang2022unsupervised,tanwisuth2023pouf} on the Domain-Net \cite{peng2019moment} and Office-Home \cite{venkateswara2017deep}.

On top of our effective training-free strategy, we propose a training-base approach (TFUP-T) to further boost VLMs' unsupervised adaptation.
Following the standard Parameter-Efficient Fine-tuning (PEFT) methods \cite{gao2021clip,zhang2021tip}, we also append the image and text adapters to the image and text encoders, respectively.
In addition, we adopt a residual connection \cite{he2016deep} to combine pre-trained features with the fine-tuned features to preserve the CLIP's generalization ability better.
By leveraging our TFUP to produce supervisory information on downstream unlabeled datasets, we can effectively tune the adaptors to achieve higher performance.
Different from existing pseudo-labeling strategies in unsupervised tuning, which mainly focus on instance-level predictions and may introduce accumulated errors \cite{tanwisuth2023pouf,zhao2023entropy}, we further introduce a marginal distribution entropy loss to constrain the model from a global perspective.
As shown in Fig. \ref{FIG:2}, our TFUP-T achieves state-of-the-art classification performance among multiple benchmarks. In particular, TFUP-T not only achieves an average accuracy improvement of 3.3\% compared to the SOTA POUF of unsupervised methods, but also obtains improvement by 1.2\% compared to KgCoOp of few-shot approaches on Domain-Net \cite{peng2019moment}.
Our contributions are summarized as follows,
\begin{itemize}
    \item We propose the first training-free approach (TFUP) for unsupervised prompt, which maximally preserves the inherent representation capabilities and enhances them with a residual connection to similarity-based prediction probabilities in a training-free and labeling-free manner.
    In particular, we generate similarity-base prediction probabilities by the Feature Cache Model and Multi-level Similarity Measure. 
    \item Based on TFUP, we propose a training-based approach (TFUP-T) to further boost performance. Considering the lack of labeled downstream data, we simultaneously optimize individual and global predictions on unlabeled data via pseudo-label cross-entropy loss and marginal distribution entropy loss.
    \item Through extensive empirical analysis, our TFUP outperforms the original CLIP on all classification datasets by a large margin. It achieves promising performance without any labeled data or training, even surpassing the training-base method on multiple classification datasets. In addition, the training-base TFUP-T obtains the new state-of-the-art compared with both unsupervised and few-shot prompt learning methods.
\end{itemize}

\section{Related Work}

\subsection{Vision-Language Models}
Large-scale vision-language models have exhibited remarkable representation capabilities on various downstream vision tasks \cite{chen2020uniter,jia2021scaling,dou2022empirical}.
By employing the contrastive learning objective to train large-scale multi-level models on 400 million image-text pairs, CLIP~\cite{radford2021learning} achieves impressive results in zero-shot visual recognition tasks. 
The success of CLIP inspired the development of several subsequent variants.
FLIP \cite{li2023scaling} randomly masks out and removes a large portion of image patches during training to increase training speed without sacrificing accuracy. 
DeCLIP \cite{li2021supervision} uses more expansive and scalable supervision to learn visual representations. 
To reduce the effect of noise in web-crawled data, SoftCLIP ~\cite{gao2023softclip} provides a soft cross-modal alignment by introducing a softened target, which is produced from the fine-grained intra-modal self-similarity.
While CLIP and its variant models achieve great results in downstream datasets, it is important to reactivate specific capabilities. 
In this work, we focus on parameter-efficient fine-tuning of large-scale vision-language models to adapt to downstream tasks.

\subsection{Prompt Learning in VLMs}
Prompt learning has become the most popular paradigm in the natural language processing community for adapting LLMs to downstream tasks \cite{lester2021power,li2021prefix,hu2021lora}.
Motivated by these studies, CoOp \cite{zhou2022learning} is the first approach which applies prompt learning to VLMs adaptation in the computer vision community.
To further enhance generalization capabilities, CoCoOp \cite{zhou2022conditional} generates image-conditional context prompts for each image and incorporates them into text prompts for prompt tuning.
In addition, KgCoOp \cite{yao2023visual} reduces the forgetting of general knowledge by minimizing the difference between text embeddings generated by learned prompts and hand-crafted prompts.
Unlike the approaches mentioned above, which train learnable prompts on labeled data of specific tasks. 
The unsupervised prompt tuning framework proposes to fine-tune models or prompts directly on unlabeled data to efficiently adapt VLMs for downstream tasks.
UPL \cite{huang2022unsupervised} selects the top-K confidence samples per class to train itself using pseudo labels generated by a pre-trained vision-language model.
POUF \cite{tanwisuth2023pouf} treats the representation of class-specific text prompts as class prototypes aligned with target image features in the latent space.
However, inaccurate pseudo-labels easily misguide the tuning process and result in poor generalization capabilities. We propose Training-Free Unsupervised Prompt (TFUP), which maximally preserves the inherent representation capabilities and enhances them while adapting VLMs to downstream tasks in a training-free and labeling-free manner.

\begin{figure*}[t]
	\centering
	\includegraphics[width=0.95\textwidth]{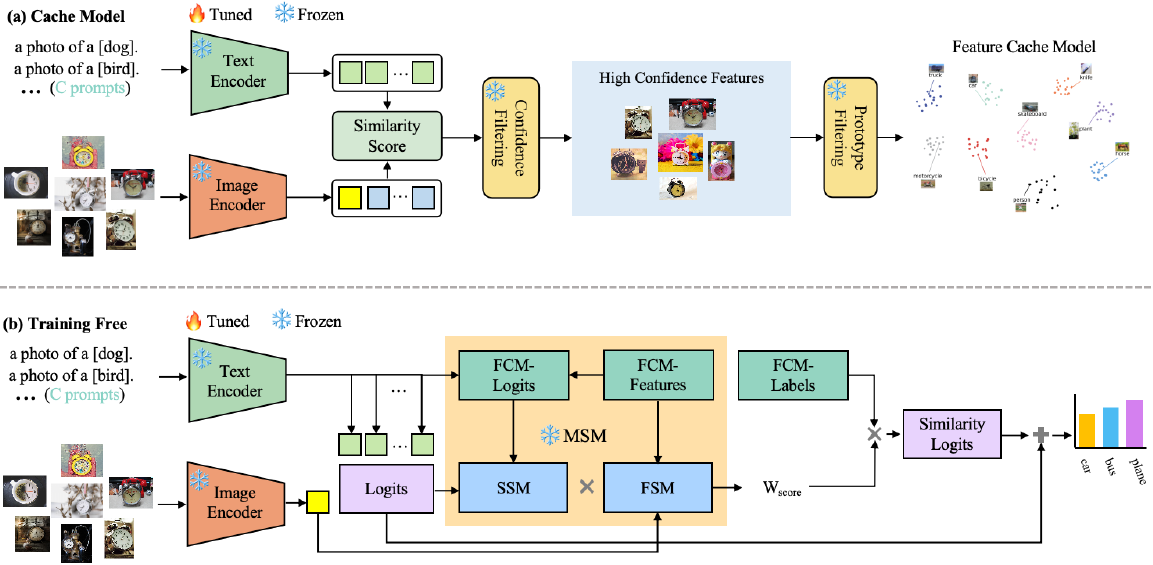}
	\caption{Overview of the TFUP framework. Our TFUP creates a Feature Cache Model (FCM) from the unsupervised training set by confidence and prototype filters. Based on the cache model, we propose a Multi-level Similarity Measure (MSM) consisting of Feature Similarity Measure (FSM) and Semantic Similarity Measure (SSM) to calculate the distance between each test image and the cached sample as the weights of corresponding cache label to generate similarity-base prediction probabilities. 
	}
	\label{FIG:arc1}
\end{figure*}

\subsection{Pseudo Labeling}
Pseudo labeling is originally proposed for semi-supervised learning and gains popularity in other domains including NLP \cite{mcclosky2006effective,he2019revisiting}, speech recognition \cite{kahn2020self,parthasarathi2019lessons},
image classification \cite{sohn2020fixmatch,zhao2022lassl},
semantic segmentation \cite{zhao2023augmentation,zhao2023instance}, object detection \cite{sohn2020simple,xu2021end}, domain adaptation \cite{choi2019pseudo}, to name a few.
The main idea is to select the class with the maximum predicted probability as pseudo labels to fine-tune the model together with the true labels.
Pseudo-Label~\cite{lee2013pseudo} is the first pseudo-label method proposed in semi-supervised learning, which selects the category with the maximum predicted probability and converts it into a hard label.
\cite{rosenberg2005semi} uses a confidence-based strategy to further filter out unlabeled samples.
Recently, as a representative work in semi-supervision learning, 
FixMatch \cite{sohn2020fixmatch} continues to employ a confidence-based strategy, retaining pseudo-labels with high-confidence predictions.
Inspired by this, our proposed TFUP generates pseudo labels for downstream data based on the original and similarity-base prediction probability. 
Benefiting from the powerful generalization performance of VLMs and  high-confidence feature cache model, 
we can effectively select the more convincing pseudo-labels.
However, these pseudo-labels methods focus mainly on individual prediction can inevitably introduce prediction bias
~\cite{zhao2023entropy}. 
Inspired by recent studies in mutual-information maximization \cite{krause2010discriminative, shi2012information, liang2020we},  we further introduce a marginal distribution entropy loss to constrain the model from a global perspective.


\section{Method}
Fig. \ref{FIG:arc1} presents an overview of our proposed TFUP. 
In this section, we first discuss a representative vision-language model, CLIP, which utilizes hand-crafted prompts in a zero-shot manner for downstream tasks in Sec. \ref{sec:3.1}.
Subsequently, we introduce our proposed 
Training-free Unsupervised Prompt (TFUP)  which maximally preserves the representational
capabilities of pre-trained VLMs while adapting them to downstream tasks in a training-free and labeling-free manner in Sec. \ref{sec:3.2}.
To further improve the performance, we then introduce an unsupervised prompt tuning method (TFUP-T) which simultaneously optimize individual and global predictions on unlabeled data via pseudo-label cross-entropy loss and marginal distribution entropy loss in Sec. \ref{sec:3.3}. 

\subsection{Preliminaries of CLIP}
\label{sec:3.1}
\paragraph{Contrastive language-image pre-training} CLIP is trained on 400 million image-text pairs with a language-image contrastive loss function. Specifically, the structure of CLIP consists of two components: visual encoder, denoted as $\mathrm{F}(\cdot)$, for converting the input images into visual features and text encoder, denoted as $\mathrm{G}(\cdot)$, for transforming input texts into text representations.
Then image and text features are projected into the same embedding space through joint training. 
After pre-training on a large dataset, it has excellent classification performance in zero-shot scenarios, demonstrating the vision-language model's excellent understanding of open-set concepts.
Consider an image classification task that 
is defined as classifying a given test image $\mathrm{x_{test}}$ into one of $\mathrm{C}$ categories.
Since the text description used in pre-training is different from the labels of the downstream recognition tasks,
CLIP places all category names into the
``\texttt{[CLASS]}''  token of a pre-defined textual template such as ``\texttt{a photo of a [CLASS]}''. We denote the text prompts obtained by converting the original labels as $\mathrm{P_{text}}$.
Then, we obtain $d$-dimensional visual features $\mathrm{f_{test}} \in \mathbb{R}^{1 \times d}$ and text features $\mathbf{F}_{\rm text} \in \mathbb{R}^{C \times d}$ by
\begin{align}
\mathrm{f_{test}}&=\mathrm{F}( \mathrm{x}_\mathrm{test}), \\
\mathbf{F}_{\rm text}&=\mathrm{G}( \mathrm{P_{text}}),
\end{align}
where $\mathrm{f}_\mathrm{test}$ and $\mathbf{F}_{\rm text}$ are L2-normalized visual and text features, respectively.
We can then obtain the classification $\mathrm{logits}\in \mathbb{R}^{1 \times C}$ by calculating the cosine similarity of visual and textual features,
\begin{equation}
\mathrm{logits}=\operatorname{softmax}(\mathrm{f}_\mathrm{test} \mathbf{F}_{\rm text}^{\top}),
\label{eq:prob}
\end{equation}
where $\mathrm{logits}$ denote the prediction probabilities for the $\rm C$ categories. We can easily identify the prediction $\hat{y} = \underset{c}{\operatorname{argmax}}\, (\mathrm{logits})$.

\subsection{Training-Free Unsupervised Prompt}
\label{sec:3.2}
We propose a training-free and labeling-free method called TFUP, which performs comparably or even better than 
training-base unsupervised methods
\cite{huang2022unsupervised,tanwisuth2023pouf}.
To achieve this goal, we construct a new Feature Cache Model (FCM) which 
stores the features of the representative sample and the corresponding one-hot labels as key-value pairs.
Based on the constructed cache model, we calculate the similarity between test image and representative samples as the weights of the corresponding labels. In addition, we design a new Multi-level Similarity Measure (MSM) method which considers both feature-level and semantic-level similarity between images.
\begin{figure*}[t]
	\centering
	\includegraphics[width=0.95\textwidth]{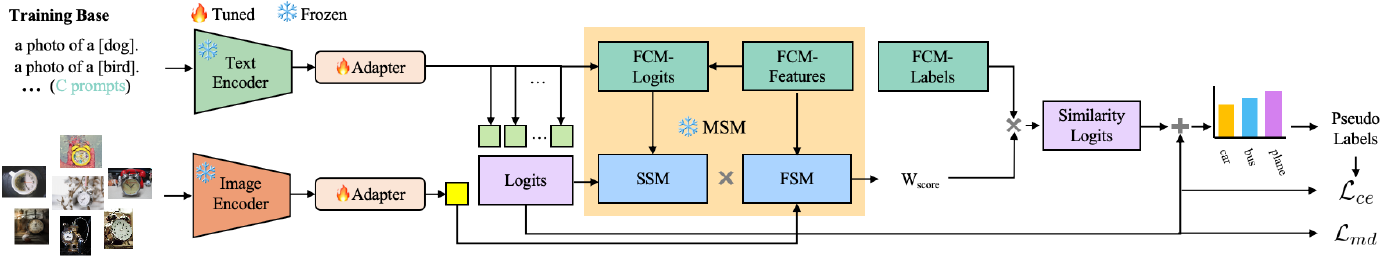}
	\caption{Framework of our proposed unsupervised prompt tuning (TFUP-T). Our TFUP-T appends CLIP model with an adapter of two-layer Multi-layer Perceptron which is optimized by the cross-entropy loss and marginal distribution entropy loss.
	}
	\label{FIG:arc2}
\end{figure*}
\paragraph{Cache model construction.}
Given the pre-trained CLIP~\cite{radford2021learning} model, we aim to leverage the unlabeled training set, denoted as $\mathrm{M}$, for unsupervised classification. 
For each training image, denoted as $\mathrm{x_i}$,
we utilize the CLIP's visual encoder to extract its $d$-dimensional visual feature, denoted as $\mathrm{f}_{\rm i} = \mathrm{F}(\mathrm{x_i})$. 
Then, we can use Eq. (\ref{eq:prob}) to generate the prediction  probability, denoted as $\mathrm{logits_i}$. Finally, we view CLIP outputs the maximum one as the prediction and the corresponding index as the pseudo label $\mathrm{L_i}= \mathrm{One Hot}(\underset{c}{\operatorname{argmax}}\, (\mathrm{logits_i}))$.
For all training samples, we denote their visual features and corresponding pseudo-labels as $\mathbf{F}_{\rm train} \in \mathbb{R}^{M \times d}$ and $\mathbf{L}_{\rm train} \in \mathbb{R}^{M \times C}$.
To create the key-value cache, we view visual features $\mathbf{F}_{\rm train}$ as keys and the corresponding pseudo-labels $\mathbf{L}_{\rm train}$ as values.
Therefore, the unsupervised dataset $\mathrm{M}$ is transformed into a key-value cache database $\mathbf{S}_{\rm train}=\langle\mathbf{F}_{\rm train}, \mathbf{L}_{\rm train}\rangle$.

For creating a standard Feature Cache Model (FCM), we first employ a confidence filter to filter out high-confidence samples and their corresponding pseudo-labels from the unsupervised dataset $\mathbf{S}_{\rm train}=\langle\mathbf{F}_{\rm train}, \mathbf{L}_{\rm train}\rangle$. 
Specifically, we select top-K most  confident samples for each class, instead of keeping all samples with confidence scores higher than a pre-defined threshold, to preserve a balanced distribution of pseudo-labeled data. Then, we denote the high confidence visual features and corresponding pseudo-labels as $\mathbf{F}_{\rm confi} \in \mathbb{R}^{K \times d}$ and $\mathbf{L}_{\rm confi} \in \mathbb{R}^{K \times C}$.
Consequently, we obtain the high confidence pseudo-label dataset $\mathbf{S}_{\rm confi}=\langle\mathbf{F}_{\rm confi}, \mathbf{L}_{\rm confi}\rangle$.

However, these high-confidence images are not necessarily representative samples.
To further refine the confident pseudo-label data into representative samples, we propose a prototype filter which calculates the cosine similarity between
each confidence sample $\mathrm{f_i} \in \mathbf{F}_{\rm confi}$ and other confidence samples by
\begin{equation}
\mathrm{S_i}={\sum_{\mathrm{j=1}}^\mathrm{K}  \cos \left(\mathrm{f_i}, \mathrm{f_j}\right)},
\end{equation}
where $\mathrm{S_i}$ denotes the prototype score of the $\mathrm{i}$-th image. 
Higher value of $\mathrm{S_i}$ means higher potential to be category prototypes. 
We select the $N$ highest scoring samples for each category as prototype samples.
Then, we obtain the prototype visual features and corresponding labels as $\mathbf{F}_{\rm proto} \in \mathbb{R}^{N \times d}$ and $\mathbf{L}_{\rm proto} \in \mathbb{R}^{N \times C}$.
Finally, we denote the prototype pseudo-label dataset as $\mathbf{S}_{\rm proto}=\langle\mathbf{F}_{\rm proto}, \mathbf{L}_{\rm proto}\rangle$.

\paragraph{Training-free inference.}
Based on the constructed feature cache model, we calculate the similarity
between test image and prototype samples as the weights of the
corresponding labels. We then combine different sample labels into a similarity-base prediction probability according to the weights.
The current similarity measure only considers the degree of similarity at the feature level.
Differently, we propose a Multi-level Similarity Measure (MSM) which considers both feature-level and semantic-level similarity to measure the similarity between test samples and prototype samples.
Specifically, for each test image, denoted as $\mathrm{x_{test}}$,
we utilize the CLIP's visual encoder to extract its $d$-dimensional visual feature, denote as $\mathrm{f}_{\rm test} = \mathrm{F}(\mathrm{x_{\rm test}})$.
Then, we calculate the multi-level similarity between test image and prototype samples as the weights of the corresponding labels.
The MSM consists of Feature Similarity Measure (FSM) and Semantic Similarity Measure (SSM).
For the FSM, we calculate the cosine similarity between
the test image features $\mathrm{f}_\mathrm{test} \in \mathbb{R}^{1 \times d}$ and the prototype features  $\mathbf{F}_{\rm proto} \in \mathbb{R}^{N \times d}$ by
\begin{equation}
\mathrm{W_{cont}}=\mathrm{f}_\mathrm{test} \mathbf{F}_{\rm proto}^{\top},
\end{equation}
where $\mathrm{W_{cont}}$ represents the feature similarity score.
In addition, the feature-level similarity measure only considers the similarity of the overall information of the image and ignores the consistency of the sample categories.
To this end, we introduce SSM which calculates the KL-divergence between
the test image prediction probabilities, denoted as $\mathrm{logits}_\mathrm{test} \in \mathbb{R}^{1 \times C}$, and the prototype prediction probabilities, denoted as $\mathrm{logits}_{\rm proto} \in \mathbb{R}^{N \times C}$, by
\begin{align}
\mathrm{\mathrm{logits}_\mathrm{test}}&=\operatorname{softmax}(\mathrm{f}_\mathrm{test} \mathbf{F}_{\rm text}^{\top}), \label{eq:Logits_test} \\
\mathrm{\mathrm{logits}_{\rm proto}}&=\operatorname{softmax}(\mathbf{F}_{\rm proto} \mathbf{F}_{\rm text}^{\top}),\\
\mathrm{W_{sem}}&=1 - \operatorname{softmax}({\mathrm{KL}(\mathrm{logits}_\mathrm{test}, \mathrm{logits}_{\rm proto})}),
\end{align}
where $\mathrm{W_{sem}}$ represents the semantic similarity score and $\mathbf{F}_{\rm text}$ denotes textual features.
$\mathrm{KL}$ computes the KL-divergence, \textit{i.e.}, $\mathrm{KL}(P,Q)=\sum_i P_i \log \frac{P_i}{Q_i}$.
To emphasize the criticality of the two measures, we combine CSM and SSM by
\begin{equation}
\mathrm{W_{fsm}}= \mathrm{W_{cont}} \circ \mathrm{W_{sem}} ,
\end{equation}
where $\circ$ denotes hadamard product and
$\mathrm{W_{fsm}}$ represents the ultimate multi-level similarity score.
Then, we combine different prototype sample pseudo-labels $\mathbf{L}_{\rm proto}$ into a similarity-base prediction probability according to the multi-level similarity scores $\mathrm{W_{fsm}}$ by
\begin{equation}
\mathrm{logits}_{\rm sim}=\mathrm{W_{fsm}} \cdot \mathbf{L}_{\rm proto},
\end{equation}
where $\cdot$ denotes matmul product and $\mathrm{logits}_{\rm sim}$ denotes the predicted probability based on feature cache model. 
After that, we obtain the final prediction probability of the test image, denote as $\mathrm{logits}_{\rm all}$, by adding the similarity-base prediction probability $\mathrm{logits}_{\rm sim}$ to the original prediction probability $\mathrm{\mathrm{logits}_\mathrm{test}}$,
\begin{equation}
\mathrm{logits_{all}} = \mathrm{\mathrm{logits}_\mathrm{test}} + \mathrm{logits}_{\rm sim},
\label{eq:Logits_all}
\end{equation}
where $\mathrm{logits_{all}}$ presents the ultimate prediction probability.

\subsection{Training-Base Unsupervised Prompt Tuning} 
\label{sec:3.3}

\paragraph{Training-base adapters}
The performance of TFUP can be further boosted by parameter efficient fine-tuning. Following the standard Parameter-Efficient Fine-tuning (PEFT) methods \cite{gao2021clip,zhang2021tip}, we attach image and text adapters to the image and text encoders, denoted by $\mathrm{T_x(\cdot)}$ and $\mathrm{T_t(\cdot)}$, respectively. Each adapter consists of two layers of linear transformations.
Referring to ReZero~\cite{bachlechner2020rezero}, we further employ two constant values $\alpha$ and $\beta$ as residual ratio to adjust the proportion of downstream knowledge and model inherent knowledge to enhance the robustness of the model and prevent over-fitting.
Mathematically, the new knowledge captured via fine-tuning is added with the original features via residual connections,
\begin{align}
    \mathrm{f_x}^{\star} &= \alpha \mathrm{T_{x}}(\mathrm{f_x}) + (1-\alpha) \mathrm{f_x}, \label{eq:alpha} \\
    \mathrm{\mathbf{F}_{\rm text}}^{\star} &= \beta \mathrm{T_{t}}(\mathbf{F}_{\rm text}) + (1 - \beta) \mathrm{\mathbf{F}_{\rm text}},
    \label{eq:beta}
\end{align}
where $\mathrm{f_x}^{\star}$ and $\mathrm{\mathbf{F}_{\rm text}}^{\star}$ denote the merged image and text features, respectively.
Then, we can use Eq. (\ref{eq:Logits_test}) to generate the prediction probabilities of the test image, denote as $\mathrm{Logits}_\mathrm{test} \in \mathbb{R}^{1 \times C}$. 

\begin{table*}[t!]
\centering
 \caption{\small Accuracy $(\%)$ on the Domain-Net \cite{peng2019moment} for CLIP based on training-free unsupervised prompt methods.}
 \setlength{\tabcolsep}{5.4mm}{
 \resizebox{1.0\textwidth}{!}{
 \begin{tabular}{c|c|cccccc|c}
\toprule  Category & Methods & \multicolumn{7}{c}{Domain-Net}  \\
\cmidrule(r){3-9} 
\multicolumn{1}{c}{} & \multicolumn{1}{|c}{} & \multicolumn{1}{|c}{ Clipart} &\multicolumn{1}{c}{ Infograph}& \multicolumn{1}{c}{ Painting}& \multicolumn{1}{c}{ Quickdraw}& 
\multicolumn{1}{c}{ Real}& 
\multicolumn{1}{c}{ Sketch}& 
\multicolumn{1}{|c}{Avg}\\
\midrule
& CLIP (Zero-Shot) \cite{radford2021learning} &  70.9 & 48.2 & 65.9 & 14.0 & 83.6 & 63.6 & 57.7 \\
& Tent \cite{wang2020tent} &  71.4 & 47.8  & 66.2 & 14.2 & 83.9 & 64.1 & 57.9\\
Unsupervised & UPL \cite{huang2022unsupervised} &  71.7 & 47.5  & 66.3 & 14.4 & 83.8 & 64.3 & 58.0\\
& POUF \cite{tanwisuth2023pouf} &  72.8 & 53.1 & 68.6 & 15.9 & 84.4 & 66.2 & 60.2 \\
\cmidrule(r){2-9} 
& {TFUP}\ (Training-Free)  & 73.9& 52.9& 69.2& 17.8& 85.2& 66.1& 60.9\\
& {TFUP-T}\ (Training-Base)  & \bf{76.0}& \bf{54.7}& \bf{72.1}& \bf{24.6}& \bf{85.8}& \bf{67.9}& \bf{63.5} \\
\midrule
Few-Shot 
& CoCoOp \cite{zhou2022conditional} & 75.1 & 55.5 & 71.5 & 20.4 & 84.8 & 67.3 & 62.4  \\
& KgCoOp \cite{yao2023visual}& 75.3 & 55.4 & 71.3 & 19.2 & 85.6 & 66.9 & 62.3  \\
\bottomrule
\end{tabular}
}
}
\label{tab:domainnet}
\end{table*}

\begin{table*}[t!]
\centering
 \caption{\small Accuracy $(\%)$ on the Office-Home  \cite{venkateswara2017deep} for CLIP based on training-free unsupervised prompt methods. }
 \setlength{\tabcolsep}{9.4mm}{
 \resizebox{1.0\textwidth}{!}{
 \begin{tabular}{c|c|cccc|c}
\toprule  Category & Methods & \multicolumn{5}{c}{Office-Home}  \\
\cmidrule(r){3-7} 
\multicolumn{1}{c}{} & \multicolumn{1}{|c}{} & \multicolumn{1}{|c}{ Art} &\multicolumn{1}{c}{ Clipart}& \multicolumn{1}{c}{ Product}& \multicolumn{1}{c}{ Real World}& 
\multicolumn{1}{|c}{Avg}\\
\midrule
& CLIP (Zero-Shot) \cite{radford2021learning} &  82.7 & 68.1 & 89.1 & 89.8 & 82.4 \\
& Tent \cite{wang2020tent} &  83.2 & 67.8 & 91.9 & 90.4 & 83.3\\
Unsupervised & UPL \cite{huang2022unsupervised} &  83.3 & 67.7  & 91.5 & 90.7 & 83.3\\
& POUF \cite{tanwisuth2023pouf} &  83.7 & 71.2 & 91.4 & 90.8 & 84.3\\
\cmidrule(r){2-7} 
& {TFUP}\ (Training-Free)  & 83.7& 71.5& 92.7& 90.6& 84.6\\
& {TFUP-T}\ (Training-Base)  & \bf{86.0}& \bf{74.2}& \bf{93.1}& \bf{91.7}& \bf{86.3} \\
\midrule
Few-Shot 
& CoCoOp \cite{zhou2022conditional} & 85.1 & 73.0 & 92.9 & 90.8 & 85.5  \\
& KgCoOp \cite{yao2023visual} & 85.0 & 73.2 & 92.7 & 91.5 & 85.6  \\
\bottomrule
\end{tabular}
}
}
\label{tab:officehome}
\end{table*}

\paragraph{Training-base supervision}  In the paradigm of unsupervised prompt tuning, it is crucial to seek appropriate supervision to train adapters on unlabeled data. Thanks to the effectiveness of our training-free strategy, we first generate pseudo-labels on unlabeled data via our TFUP. Thus we can utilize Eq. (\ref{eq:Logits_all}) to generate the prediction probabilities of the unlabeled instances, denote as $\mathrm{Logits}_\mathrm{all} \in \mathbb{R}^{1 \times C}$. We then view the maximum output of the CLIP model as the prediction and the corresponding index as the pseudo label $\mathrm{L}_{\mathrm{all}}= \mathrm{One Hot}(\underset{c}{\operatorname{argmax}}\, (\mathrm{logits}_{\mathrm{all}}))$. After that, we train the adapters via a standard cross-entropy loss,
\begin{equation}
\mathcal{L}_{ce} = {\mathbf{1}}\left(\max \left(\mathrm{Logits}_\mathrm{test}\right) \geq \theta\right) \mathrm{CE}\left(\mathrm{L}_\mathrm{all}, \mathrm{Logits}_\mathrm{test} \right),
\label{eq:ce}
\end{equation}
where  $\mathbf{1}$ denotes retaining only the high-confidence predictions higher than pre-defined threshold~$\theta$. 
However, all these pseudo-label-based methods~\cite{lee2013pseudo,mclachlan1975iterative} focus solely on the instance-level constraint, ignoring the significance of the global prediction statistic on unlabeled data. Though high-confidence filtering can effectively select the more convincing pseudo-labels,  constraining only on individual prediction can inevitably introduce prediction bias~\cite{zhao2023entropy}, considering the different learning difficulties of distinct classes, especially with a huge category space. To this end, inspired by recent studies in mutual-information maximization \cite{krause2010discriminative, shi2012information, liang2020we},  we further introduce a marginal distribution entropy loss to constrain the model from a global perspective,
\begin{equation}
    \mathcal{L}_{md} = {\log \rm{C}}- [-\sum_{c} \mathrm{h}_c \log \mathrm{h}_c],
\end{equation}
where $\log \rm{C}$ is the maximum value of distribution entropy and $\mathrm{h}_c$ denotes the marginal distribution of the prediction probabilities of the test images on the class index $c, c\in\{0,1,...,C-1\}$ .

\section{Experiments}
In this section, we evaluate the performance of the training-free TFUP and the training-base TFUP-T on four public datasets including 
Domain-Net \cite{peng2019moment}, 
Office-Home \cite{venkateswara2017deep},
Office-31 \cite{saenko2010adapting},
and VisDA-2017 \cite{peng2017visda}. 
We compare our methods with recent prompt learning methods including Clip (Zero-Shot) 
\cite{radford2021learning}, Tent \cite{wang2020tent}, UPL \cite{huang2022unsupervised}, POUF \cite{tanwisuth2023pouf}, CoCoOp \cite{zhou2022conditional}, and KgCoOp \cite{yao2023visual}.
Additionally, we provide extensive experimental results and further ablation studies. 

\subsection{Datasets}
{1)} {\it Domain-Net \cite{peng2019moment} } is the largest and the most challenging domain adaptation benchmark which contains approximately 600,000 images from 6 different domains including Clipart (clp), Infograph (inf), Painting
(pnt), Quickdraw (qdr), Real (rel) and Sketch (skt).
{2)} {\it Office-Home \cite{venkateswara2017deep} } refers to a difficult domain adaptation dataset which includes 15500 images in office and home extracted from 4 different domains: Art (A), Clip (C), Product
(P) and RealWorld (R). 
{3)} {\it Office-31~\cite{saenko2010adapting} } is a standard dataset in visual transfer learning which contains 4,652 images with 31 classes from three domains: Amazon (A), Webcam (W), and DSLR (D).
{4)} {\it VisDA-2017 \cite{peng2017visda} } is a dataset from the 2017 Vision Domain Adaptation Challenge, covering 12 categories and more than 280,000 images, including 152,397 synthetic images and 55,388 real images.

\begin{table*}[t!]
\centering
  \caption{\small Accuracy $(\%)$ on the Office-31 \cite{saenko2010adapting} for CLIP based on training-free unsupervised prompt methods. }
 \setlength{\tabcolsep}{12.4mm}{
 \resizebox{1.0\textwidth}{!}{
 \begin{tabular}{c|c|ccc|c}
\toprule  Category & Methods & \multicolumn{4}{c}{Office-31}  \\
\cmidrule(r){3-6} 
\multicolumn{1}{c}{} & \multicolumn{1}{|c}{} & \multicolumn{1}{|c}{ Amazon} &\multicolumn{1}{c}{ Dslr}& \multicolumn{1}{c}{ Webcam}&
\multicolumn{1}{|c}{Avg}\\
\midrule
& CLIP (Zero-Shot) \cite{radford2021learning} &  79.0 & 77.5 & 74.7 & 77.1 \\
& Tent \cite{wang2020tent} &  81.5 & 80.7 & 82.8 & 81.7\\
Unsupervised & UPL \cite{huang2022unsupervised} &  81.4 & 82.6  & 83.6 & 82.5\\
& POUF \cite{tanwisuth2023pouf}  & 83.6 & 89.9 & 90.6 & 88.0 \\
\cmidrule(r){2-6} 
& {TFUP}\ (Training-Free)  & 81.2& 86.9& 84.2& 84.1 \\
& {TFUP-T}\ (Training-Base)  & \bf{84.8}& \bf{90.8}& \bf{93.2}& \bf{89.6} \\
\midrule
Few-Shot 
& CoCoOp \cite{zhou2022conditional} & 83.9 & 92.4 & 94.2 & 90.2  \\
& KgCoOp \cite{yao2023visual} & 84.4 & 92.6 & 94.2 & 90.4  \\
\bottomrule
\end{tabular}
}
}
\label{tab:office31}
\end{table*}

\begin{table*}[t!]
\centering
 \caption{\small Accuracy $(\%)$ on the VisDA-2017 \cite{peng2017visda} for CLIP based on training-free unsupervised prompt methods. }
 \setlength{\tabcolsep}{3.2mm}{
 \resizebox{1.0\textwidth}{!}{
 \begin{tabular}{c|c|cccccccccccc|c}
\toprule  Category & Methods & \multicolumn{13}{c}{VisDA-2017}  \\
\cmidrule(r){3-15} 
\multicolumn{1}{c}{} & \multicolumn{1}{|c}{} & \multicolumn{1}{|c}{ plane} &\multicolumn{1}{c}{ bicycle}& \multicolumn{1}{c}{ bus}& \multicolumn{1}{c}{ car}& 
\multicolumn{1}{c}{ horse}& \multicolumn{1}{c}{ knife}& \multicolumn{1}{c}{ mcycl}& \multicolumn{1}{c}{ person}& \multicolumn{1}{c}{ plant}& \multicolumn{1}{c}{ sktbrd}& \multicolumn{1}{c}{ train}& \multicolumn{1}{c}{ truck}&
\multicolumn{1}{|c}{Avg}\\
\midrule
& CLIP (Zero-Shot) \cite{radford2021learning} &  99.1 & 91.7 & \bf{93.8} & 76.6 & 98.4 & 91.5 & 95.3 & 82.7 & 86.5 & 96.0 & 94.6 & 60.2 & 88.9 \\
& Tent \cite{wang2020tent} &  98.7 & 91.3  & 86.9 & \bf{89.1} & 98.1 & 94.8 & 92.9 & \bf{85.4} & 90.1 & 92.2 & 94.1 & 48.7 & 88.5\\
Unsupervised & UPL \cite{huang2022unsupervised} &  99.0 & 93.0  & 91.3 & 77.8 & 98.4 & 94.7 & 93.8 & 83.3 & 87.1 & 96.1 & 94.5 & \bf{67.5} & 89.7\\
& POUF \cite{tanwisuth2023pouf}  & 99.0& 91.0& 92.0& 80.3& 98.7& 94.7& 95.4 & 81.1 & 89.5 & \bf{96.5} & 95.2 & 64.7 & 89.8\\
\cmidrule(r){2-15} 
& {TFUP}\ (Training-Free)  & 99.1& \bf{93.7}& 91.2& 83.1& 98.0& 94.0& 92.8 & 82.7 & 87.4 & 94.9 & 94.7 & 63.5 & 89.6\\
& {TFUP-T}\ (Training-Base)  & \bf{99.2}& 92.0& 87.3& 82.6& \bf{99.1}& \bf{96.4}& \bf{96.4} & 83.6 & \bf{92.8} & 94.2 & \bf{96.2} & 66.1 & \bf{90.5}\\
\midrule
Few-Shot 
& CoCoOp \cite{zhou2022conditional} & 99.1 & 93.6 & 91.9 & 74.6 & 98.4 & 93.4 & 91.7 & 61.3 & 87.0 & 96.8 & 95.0 & 68.7 & 87.6  \\
& KgCoOp \cite{yao2023visual} & 99.2 & 92.3 & 93.6 & 76.6 & 98.3 & 90.3 & 94.6 & 83.6 & 85.3 & 96.1 & 94.3 & 62.6 & 88.9 \\
\bottomrule
\end{tabular}
}
}
\label{tab:visda2017}
\end{table*}

\subsection{Baselines}
To verify the effectiveness of our approach, we compare it with plenty of advanced methods: (1) Zero-shot CLIP \cite{radford2021learning}, which applies hand-crafted prompts; (2) Tent \cite{wang2020tent} refers to using entropy minimization to tune the model before making predictions on downstream datasets; (3) UPL \cite{huang2022unsupervised} selects the top-K confidence samples per class to train the soft prompts using pseudo labels generated by a pre-trained vision-language model. For a fair comparison, we do not use model ensemble as done in \cite{huang2022unsupervised};
(4) POUF \cite{tanwisuth2023pouf} treats the representation of class-specific text prompts as class prototypes aligned with target image features in the latent space;
(5) CoCoOp \cite{zhou2022conditional} proposes an image-condition prompt learning method to generate specific text prompts for each image input;
(6) KgCoOp \cite{yao2023visual} reduces the difference between text features generated by learnable prompts and hand-crafted prompts to enhance the generalization ability of learnable prompts.
The reported results for the baseline models are obtained using the officially open-source code.

\subsection{Implementation Details}
Our implementation is based on the open-source repository of POUF \cite{tanwisuth2023pouf}. For all experiments, we follow the same unlabeled downstream dataset, visual and text encoder, data augmentation, learning rate schedule, and batch size. 
Differently, we set the hyper-parameter $\alpha = 0.2$ and $\beta = 0.5$ in Eq. (\ref{eq:alpha}) and Eq. (\ref{eq:beta}), which are analyzed in our experiments. 
Following previous pseudo-labeling strategy, we set the predefined threshold $\theta$ in Eq. (\ref{eq:ce}) to 0.95 as a constant value for all experiments.
The final performance reported below is the average of three runs with different random seeds. All experiments are conducted based on RTX 3060.

\subsection{Experimental Results}

\paragraph{Results on Domain-Net \cite{peng2019moment}}
Tab. \ref{tab:domainnet} reports the results comparing TFUP and TFUP-T with unsupervised prompt tuning and few-shot prompt learning methods across 6 domains
on the Domain-Net \cite{peng2019moment} datasets.
It is obvious that our TFUP achieves the new state-of-the-art (SOTA) performance without any training compared with previous unsupervised prompt tuning methods. Specifically, TFUP outperforms original CLIP with prompt engineering by 3.2\%. 
In addition to prompt tuning, TFUP-T not only achieves
an average accuracy improvement of 3.3\% compared to
the current SOTA POUF of unsupervised methods, but also obtains improvement by 1.2\% compared to the SOTA KgCoOp of  few-shot approaches.
The above experiment demonstrates the efficiency and effectiveness of TFUP, as well as the superiority of the training-base TFUP-T.

\paragraph{Results on Office-Home \cite{venkateswara2017deep}}
Tab. \ref{tab:officehome} illustrated the results comparing TFUP and TFUP-T with unsupervised prompt tuning and few-shot prompt learning methods across 4 domains
on the Office-Home \cite{venkateswara2017deep} datasets.
Our TFUP achieves new SOTA performance without any training compared with unsupervised tuning methods and outperforms original CLIP with prompt engineering by 2.2\%. 
For prompt tuning, TFUP-T boosts the performance of the POUF by 2.0\% and demonstrates clear advantages over few-shot methods.
\paragraph{Results on Office-31~\cite{saenko2010adapting}}
Tab. \ref{tab:office31} provides the results comparing TFUP and TFUP-T with unsupervised prompt tuning and few-shot prompt learning methods across 3 domains
on the Office-31~\cite{saenko2010adapting} datasets.
Our TFUP has significant advantages without any training compared with Tent and UPL. Although the accuracy of training-free TFUP is slightly lower than the trainable POUF, TFUP-T achieves new SOTA on all domains.
Due to the smaller amount of unsupervised data in the Office-31 dataset compared to the other datasets, the performance achieved by our method is limited.

\paragraph{Results on VisDA-2017 \cite{peng2017visda}}
Tab. \ref{tab:visda2017} summarizes the results compare TFUP and TFUP-T with unsupervised prompt tuning and few-shot prompt learning methods on the VisDA-2017 \cite{peng2017visda} datasets.
Obviously, the training-free TFUP demonstrates competitive performance compared to the other unsupervised methods.
By efficient fine-tuning, TFUP-T further improves the performance of model, confirming the effectiveness and versatility of our methods.

\subsection{Ablation analysis}
\paragraph{Effectiveness of each component.}
To evaluate the effectiveness of various components, we conduct ablation experiments on Office-Home and  Domain-Net datasets, as reported in Tab. \ref{tab:component}. 
We can clearly see that each component significantly improves the performance of the model.
For training-free TFUP, FCM + FSM increases the average accuracy of Office-Home and Domain-Net datasets by 2.2\% and 3.2\%, respectively. 
By efficient fine-tuning, $\mathcal{L}_{ce}$ further improves the performance of the model.
Additionally, the pseudo-labeling strategy mainly focuses on rectifying individual predictions. 
We introduce a global prediction constraint $\mathcal{L}_{md}$, which increases the average performance of CLIP
from 57.7\% to 63.5\% on the most challenging Domain-Net.
It demonstrates the effectiveness of each component, and the superiority of  our TFUP and TFUP-T.

\begin{table}[t]
  \centering
  \small
  \caption{ \textbf{Ablation studies of our method.} The average results of all domains on Office-Home and Domain-Net datasets are reported. Four ablation cases are considered: 
FCM: The feature cache model. 
FSM: The multi-level similarity measure. $\mathcal{L}_{\rm ce}$: The pseudo-label cross-entropy loss. $\mathcal{L}_{\rm md}$: The marginal distribution entropy loss.}
  \setlength{\tabcolsep}{3.5mm}{
  \resizebox{1.0\linewidth}{!}{
  \begin{tabular}{c|c|c|c|c|c}
    \toprule
    \multicolumn{4}{c|}{Component Module} & 
    \multicolumn{2}{c}{Average Accuracy} \\
    \cmidrule(lr){1-6}
    FCM & MSM & $\mathcal{L}_{\rm ce}$ & $\mathcal{L}_{\rm md}$ & Office-Home & Domain-Net  \\
    \midrule
    &    &     &   & 82.4  & 57.7 \\  
    \checkmark &   &   &   &  83.8 {(\hgreen{$+$1.4})} & 59.7 {(\hgreen{$+$2.0})} \\
    \checkmark & \checkmark &  &  & 84.6 {(\hgreen{$+$2.2})} & 60.9 {(\hgreen{$+$3.2})}\\
    \checkmark & \checkmark & \checkmark &  & 85.4 {(\hgreen{$+$3.0})}  & 62.3 {(\hgreen{$+$4.6})}\\
    \midrule
    \checkmark & \checkmark & \checkmark & \checkmark & 86.3 {(\hgreen{$+$3.9})}  & 63.5 {(\hgreen{$+$5.8})}\\
    \bottomrule
  \end{tabular}
  }
  }
  \label{tab:component}
\end{table}

\begin{table}[t]
\caption{The experiment results of different sample filter strategies on Office-Home dataset.} 
\small
    \centering
    \setlength{\tabcolsep}{1.6mm}{
    \resizebox{1.0\linewidth}{!}{
		\begin{tabular*}{\hsize}{l|c|c|c|c|c}
            \toprule
            {Filtering Strategy} & 
           \multicolumn{5}{c}{Office-Home Dataset} \\
           \cmidrule(l){2-6}
            &A  & C & P & R & Average\\
           \midrule
            Unsupervised Dataset &82.5&68.3 &90.9&89.8 &82.9\\
            \hline
            + Confidence Filter&83.4&70.3 &91.9&90.0&83.9 {(\hgreen{$+$1.0})}\\
            + Prototype Filter&83.5&70.9 &92.5&90.5&84.3 {(\hgreen{$+$1.4})}\\
            + Double Filter&83.7&71.5 &92.7&90.6&84.6 {(\hgreen{$+$1.7})}\\
			\bottomrule
	\end{tabular*}
 }
 }
	\label{tab:filter}
\end{table}

\paragraph{Different sample filter strategies.}
As presented in Tab. \ref{tab:filter}, we compare several common sample filter strategies to assess the effectiveness of our approach. 
i) Confidence Filter Strategy. Select top-K high confidence samples by pseudo labels.
ii) Prototype Filter Strategy. Select top-K representative samples by prototype score.
iii) Double Filter Strategy. Combine the confidence filter and prototype filter strategy to obtain the representative samples.
Compared with the overall unlabeled data, both of the filter strategy designs could improve the final results by large margins. 
It demonstrates that the key factor of the filter strategy to achieve significant performance is reducing the introduction additional noisy samples.
In addition, we find that combining confidence and prototype filter can both improve the results compared with using either of them. It indicates that only selecting the high confidence 
samples is not enough, and prototype scores are needed to screen representative samples.

\begin{table}[t]
\small
    \centering
    \caption{The experiment results of different similarity measure strategies on Office-Home dataset.}
    \setlength{\tabcolsep}{1.4mm}{
		\begin{tabular*}{\hsize}{l|c|c|c|c|c}
            \toprule
            {Similarity Measure} & 
           \multicolumn{5}{c}{Office-Home Dataset} \\
           \cmidrule(l){2-6}
            &A  & C & P & R & Average\\
           \midrule
            CLIP (Zero-Shot) &82.7&68.1 &89.1&89.8&82.4\\
            \hline
            + Feature Similarity &83.3&69.7 &92.0&90.2&83.8 (\hgreen{$+$1.4})\\
            + Semantic Similarity &83.6&70.9 &92.3&90.5&84.3 (\hgreen{$+$1.9})\\
            + Multi-level Similarity &83.7&71.5 &92.7&90.6&84.6 (\hgreen{$+$2.2})\\
			\bottomrule
	\end{tabular*}
    } 
	\label{tab:measure}
\end{table}

\begin{figure}[t]    
    \centering
    \includegraphics[width=0.49\textwidth]{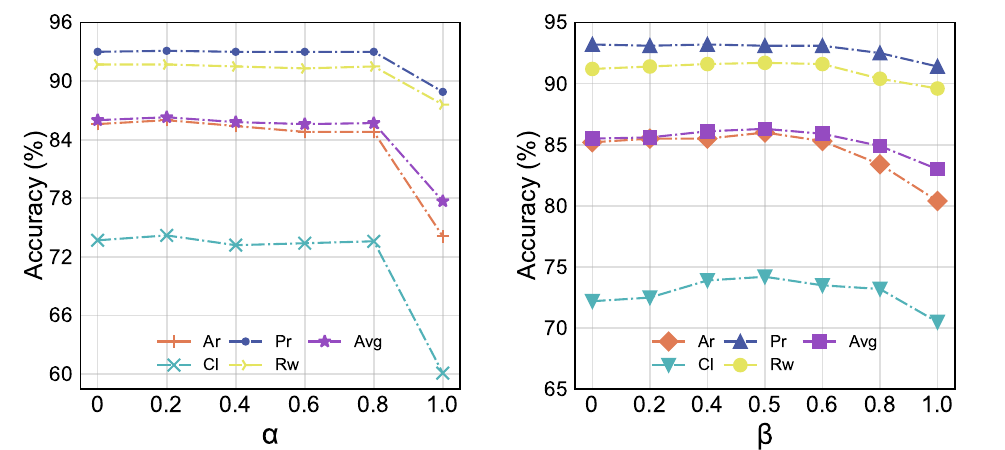}
    \setlength{\abovecaptionskip}{-0.2em}
    \caption{\textbf{Sensitivity analysis of  $\alpha$ and ${\beta}$} on Office-Home.}
  \label{FIG:AblationStudy}
\end{figure}

\paragraph{Different similarity measure strategies.}
As summarized in Tab. \ref{tab:measure},
we provide in-depth analysis about the similarity measure strategies.
i) Feature Similarity Measure (FSM) leverages cosine similarity to measure the similarity of the overall information of the images.
ii) Semantic Similarity Measure (SSM) leverages KL-divergence to measure the distance of the image prediction probabilities.
iii) Multi-level Similarity Measure (MSM) considers both feature-level and semantic-level similarities between images.
We observe that FSM and SSM improve the average accuracy by 1.4\% and 1.9\% respectively, and the combination of them improves the results by 2.2\%. 
It demonstrates that the effectiveness of considering both feature-level and semantic-level similarities, which not only measures the degree of similarity of the overall image information, but also ensures the semantic consistency of similar samples.

\paragraph{Sensitivity analysis.}
As shown in Fig. \ref{FIG:AblationStudy}, we evaluate the hyper-parameter sensitivity of  $\alpha$ and $\beta$ of Eq. (\ref{eq:alpha}) and Eq. (\ref{eq:beta}) across 4 domains on the Office-Home datasets.
$\alpha$ and $\beta$ are  residual ratio to adjust the proportion of new knowledge and inherent knowledge.
Obviously, on most domains, the best residual rate of image features is $\alpha$=0.2, and the best residual rate of text features is $\beta$=0.5.
The model performance will decrease whether the optimal parameters increase or decrease. 
It may be because too much new knowledge causes the model to overfit while too little new knowledge makes it difficult for the model to adapt to a specific task.
Therefore, we set $\alpha$=0.2 and $\beta$=0.5 to obtain the best performance trade-off.

\section{Conclusion}
In this paper, we propose a novel approach named Training-Free Unsupervised Prompt (TFUP), which maximally preserves the inherent representation capabilities and enhances them with a residual connection to similarity-based prediction probabilities in a training-free and labeling-free manner.
We generate similarity-base prediction probabilities by the proposed Feature
Cache Model (FCM) and Multi-level Similarity Measure (MSM). 
In this way, TFUP outperforms original CLIP on all classification datasets by a large margin. 
It achieves promising performance without any labeled data or training, even surpassing the training-base prompt learning methods on multiple classification datasets. 
By efficient fine-tuning, TFUP-T
not only achieves the SOTA compare with unsupervised prompt learning approaches but also demonstrates clear advantages over few-shot prompt learning methods.
We hope that this paper, which maximally preserves the inherent representation capabilities and enhances them while adapting pre-train VLMs to specific downstream tasks in a training-free and labeling-free manner, will provide insights for future work on unsupervised prompt tuning.



\end{document}